\documentclass[10pt,twocolumn,letterpaper]{article}

\usepackage[]{cvpr}      
\usepackage{multirow}

\usepackage[dvipsnames]{xcolor}

\definecolor{darkred}{rgb}{0.7, 0.0, 0.0}
\definecolor{darkgreen}{rgb}{0.0, 0.37, 0.14}
\definecolor{darkblue}{rgb}{0.10, 0.17, 0.8}

\newcommand{\abbrmethod}{HDRFlow\xspace}

\definecolor{cvprblue}{rgb}{0.21,0.49,0.74}
\usepackage[pagebackref,breaklinks,colorlinks,citecolor=cvprblue]{hyperref}
\usepackage[T1]{fontenc}

\def\paperID{810} 
\def\confName{CVPR}
\def\confYear{2024}

\title{HDRFlow: Real-Time HDR Video Reconstruction with Large Motions}

\author{
Gangwei Xu$^{1,2}$\footnotemark[1],
~~Yujin Wang$^{2}$\footnotemark[1],
~~Jinwei Gu$^{3}$,
~~Tianfan Xue$^{3}$,
~~Xin Yang$^{1}$\footnotemark[2]\\
[2mm]
$^1$~School of EIC, Huazhong University of Science and Technology \\ 
$^2$~Shanghai AI Laboratory \quad $^3$~The Chinese University of Hong Kong\\ 
{\tt\small \{gwxu, xinyang2014\}@hust.edu.cn, wangyujin@pjlab.org.cn}\\
{\tt\small \{jwgu@cse, tfxue@ie\}.cuhk.edu.hk}
}

\begin{document}
\maketitle
\begin{abstract}
Reconstructing High Dynamic Range (HDR) video from image sequences captured with alternating exposures is challenging, especially in the presence of large camera or object motion. Existing methods typically align low dynamic range sequences using optical flow or attention mechanism for deghosting. However, they often struggle to handle large complex motions and are computationally expensive. To address these challenges, we propose a robust and efficient flow estimator tailored for real-time HDR video reconstruction, named HDRFlow. HDRFlow has three novel designs: an HDR-domain alignment loss (HALoss), an efficient flow network with a multi-size large kernel (MLK), and a new HDR flow training scheme. The HALoss supervises our flow network to learn an HDR-oriented flow for accurate alignment in saturated and dark regions. The MLK can effectively model large motions at a negligible cost. In addition, we incorporate synthetic data, Sintel, into our training dataset, utilizing both its provided forward flow and backward flow generated by us to supervise our flow network, enhancing our performance in large motion regions. Extensive experiments demonstrate that our HDRFlow outperforms previous methods on standard benchmarks. To the best of our knowledge, HDRFlow is the first real-time HDR video reconstruction method for video sequences captured with alternating exposures, capable of processing 720p resolution inputs at 25ms. Project website: \textcolor{magenta}{https://openimaginglab.github.io/HDRFlow/}.
\end{abstract}   

\renewcommand{\thefootnote}{\fnsymbol{footnote}}
\footnotetext[1]{Equal contribution. This work was done when Gangwei Xu interned at Shanghai AI Laboratory.}
\footnotetext[2]{Corresponding author.}

\section{Introduction}
\label{sec:intro}

Capturing a high dynamic range (HDR) natural scene using a standard digital camera with a limited dynamic range (LDR) often yields undesirable results: either details in highlights are missing or shadows are too dark. The most prevalent solution to this issue is video HDR fusion, which merges multiple LDR images with varying exposures. With recent advances in deep learning, video HDR fusion has become a popular solution for HDR capturing and has been extensively used on recent mobile cameras.

\begin{figure}[t]
\centering
{\includegraphics[width=1.0\linewidth]{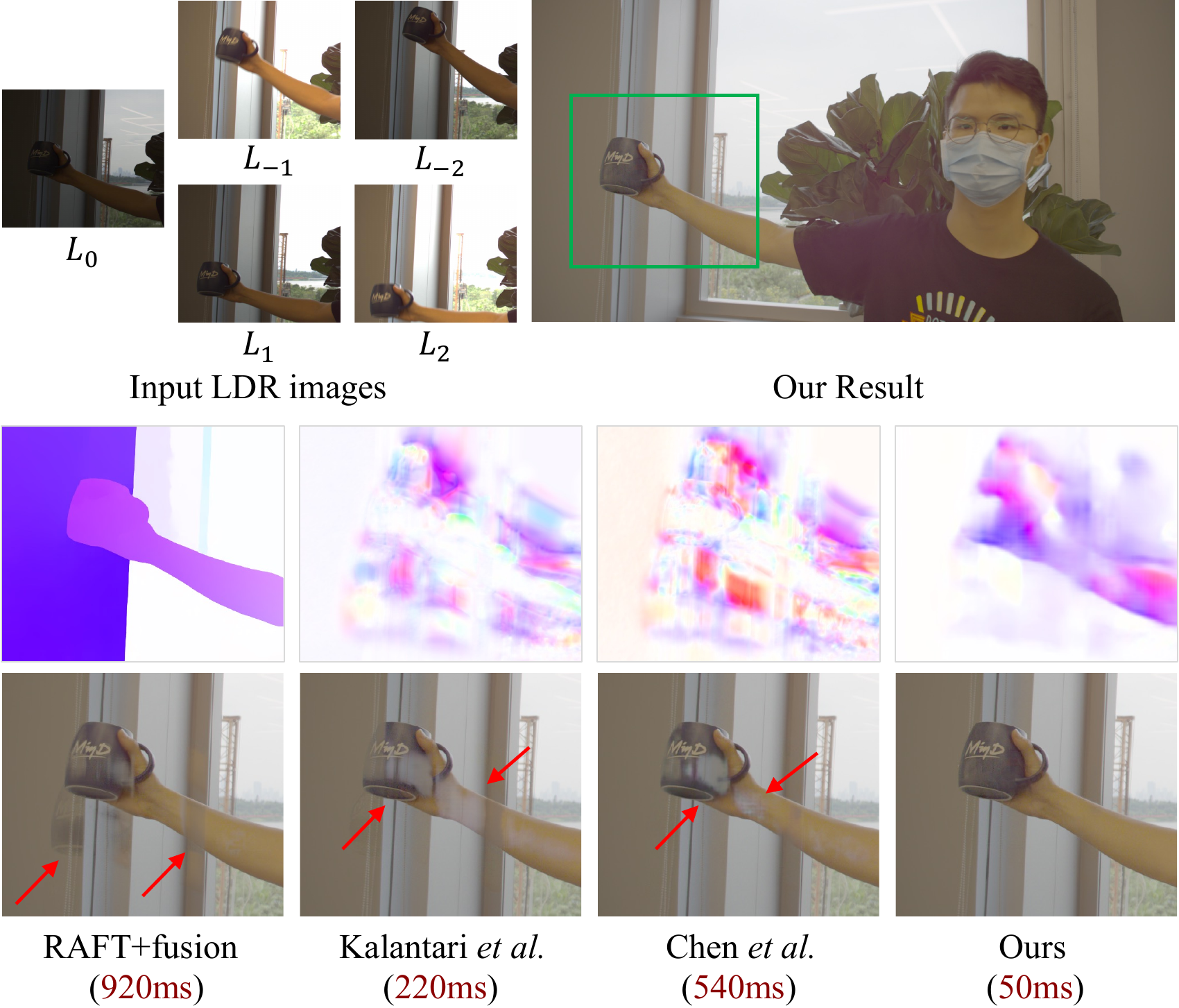}}
\caption{\textbf{Row 2:} the optical flow from frame 0 to frame -1. \textbf{Row 3:} the resulting HDR images. The methods of Kalantari \etal~\cite{kalantari19} and Chen \etal~\cite{chen2021hdr} struggle to predict accurate optical flow due to large motions, resulting in ghosting artifacts in the HDR output. In contrast, our HDRFlow predicts HDR-oriented optical flow and exhibits robustness to large motions. We compare our HDR-oriented flow with RAFT's~\cite{raft} flow. RAFT's flow is sub-optimal for HDR fusion, and alignment may fail in occluded regions. leading to significant ghosting artifacts in the HDR output.}
\label{fig:teaser}
\vspace{-15pt}
\end{figure}

\begin{figure}
\centering
{\includegraphics[width=1.0\linewidth]{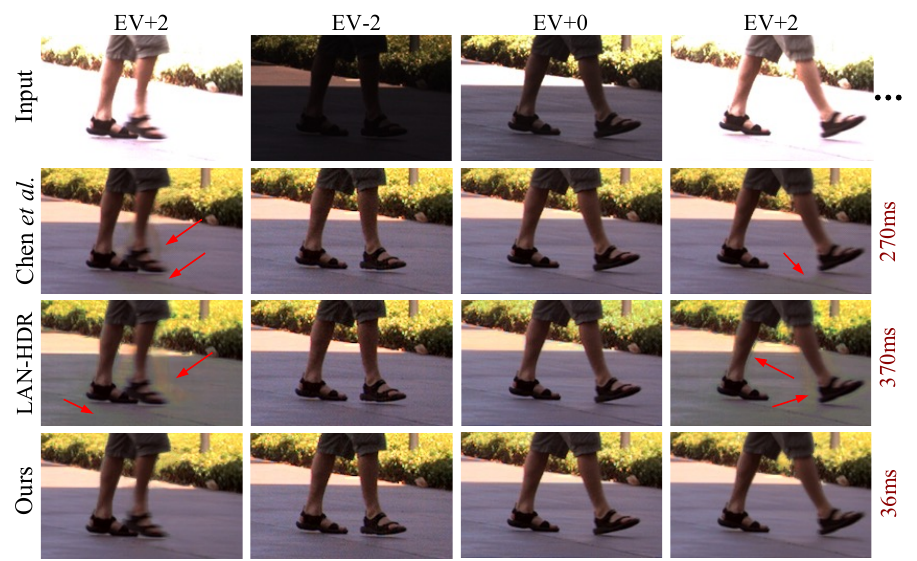}}
\caption{HDR video reconstruction from sequences (image size: $1280\!\times\!720$) captured with three alternating exposures. Row 1 displays four input LDR frames. Rows 2-4 are the reconstructed HDR frames using methods Chen \etal~\cite{chen2021hdr}, LAN-HDR~\cite{lan_hdr} and ours.}\label{fig:hdrvideo}
\vspace{-10pt}
\end{figure}

However, in the presence of large motion, video HDR fusion algorithms still face two main challenges: achieving robust artifact-free merging and ensuring efficient processing for real-time applications. Large motions are common in amateur captures, caused by either camera motion during hand-held capture, or by object motion, such as people walking. Without accurate alignment of neighboring frames, ghosting artifacts will appear in the fused HDR video, as demonstrated in \cref{fig:teaser} and \cref{fig:hdrvideo}. Researchers have attempted to enhance alignment robustness using advanced optical flow methods, such as SpyNet~\cite{spynet}, employed by Kalantari~\etal~\cite{kalantari19} and Chen~\etal~\cite{chen2021hdr}. 
However, these flow networks are not specifically designed for HDR fusion and may struggle with inputs of varying exposure. Alternatively, other researchers~\cite{lan_hdr} have attempted direct fusion of multiple frames using an attention module, but this approach is susceptible to aggregating irrelevant information due to inaccurate attention maps, leading to local inconsistencies (\cref{fig:hdrvideo}, \cref{fig:comp_att}).
Furthermore, these alignment modules tend to be computationally intensive, especially in the presence of substantial motion. Thus, off-the-shelf flow or attention modules may be sub-optimal for HDR fusion.

For robust and efficient alignment, we design a flow estimation algorithm that is tailored for real-time HDR video reconstruction, named \textit{HDRFlow}. The proposed HDRFlow has included three novel designs: an HDR-domain alignment loss, an efficient flow network with a multi-size large kernel, and a new HDR flow training scheme.

Firstly, to train robust alignment tailored for HDR fusion, we propose a novel HDR-domain alignment loss (HALoss). A simple solution to align input frames is to use a pre-trained model on extensive optical flow datasets, such as RAFT~\cite{raft}. However, RAFT's flow is sub-optimal, and alignment may fail in occluded regions (\cref{fig:teaser}). A better solution for handling occlusion and large motion is to use a task-oriented flow trained with an unsupervised approach, as illustrated by Xue~\etal~\cite{xue2019video}. A typical unsupervised flow training calculates the photometric loss between the aligned frames, but this photometric loss relies on the brightness consistency assumption between input frames, which does not hold for HDR fusion, where input frames have different exposures. Therefore, we propose a novel HDR-domain alignment loss that is robust to varying exposures in input and predicts an HDR-oriented flow for precise alignment.

Secondly, we introduce a novel alignment network using multi-size large kernel convolutions, which can efficiently handle large motions. To manage extensive movement, most existing optical flow methods utilize deep iterative structure~\cite{raft,gma,flowformer,flowformer++,xu2023memory,xu2023iterative,videoflow,pwc,xu2023accurate} or Transformer~\cite{gmflow, unifying, transflow}, capable of estimating flow with large motions, albeit at a high computational cost. In contrast, we present a simple flow network with multi-size large kernel convolutions that only operate on low-resolution input, proficient in handling large motions with minimal computational expense.
The flow boundaries may not be as sharp as those produced by computationally expensive methods (\cref{fig:teaser}), but the fused HDR image retains clarity and sharpness, showing a sharp flow boundary is not imperative for HDR fusion. 
Consequently, our flow network predicts bidirectional optical flows while requiring only 10 ms for 720p resolution inputs.

Thirdly, we introduce a new HDR flow training scheme that integrates both synthetic and real videos for video HDR training. Most existing video HDR networks are mostly trained on real videos, such as Vimeo-90K~\cite{xue2019video}, which often lack instances of large motion. 
In this work, we introduced the synthetic data, Sintel~\cite{sintel}, into our training dataset, utilizing both its provided ground-truth forward flow and backward flow generated by us to supervise our flow network.  
Compared to networks trained solely on real videos, it improves the flow’s robustness against large motions.

Extensive experiments demonstrate that our approach surpasses state-of-the-art methods~\cite{kalantari19,chen2021hdr,lan_hdr} on public benchmarks~\cite{cinematic,chen2021hdr}. Specifically, our approach exhibits superior performance in handling large motions (~\cref{fig:teaser}, \cref{fig:motion}).
In summary, our main contributions are as follows:
\begin{itemize}
    \item We introduce a novel HDR-domain alignment loss to supervise the flow network, enabling accurate alignment in saturated and dark regions.
    \item We propose a lightweight flow network with multi-size large kernel convolutions to efficiently model large motions. It predicts bidirectional optical flows with a runtime of 10ms for 720p resolution. The total HDR fusion only takes 25ms, 10$\times$ faster than the current state-of-the-art methods~\cite{chen2021hdr,lan_hdr}.
    \item We propose a novel training scheme that incorporates both synthetic and real videos for training, further boosting the robustness of our network under large motions.
\end{itemize}


\section{Related Work}
\label{sec:related}

\begin{figure*}[t]
    \centering
    \includegraphics[width=0.9\linewidth]{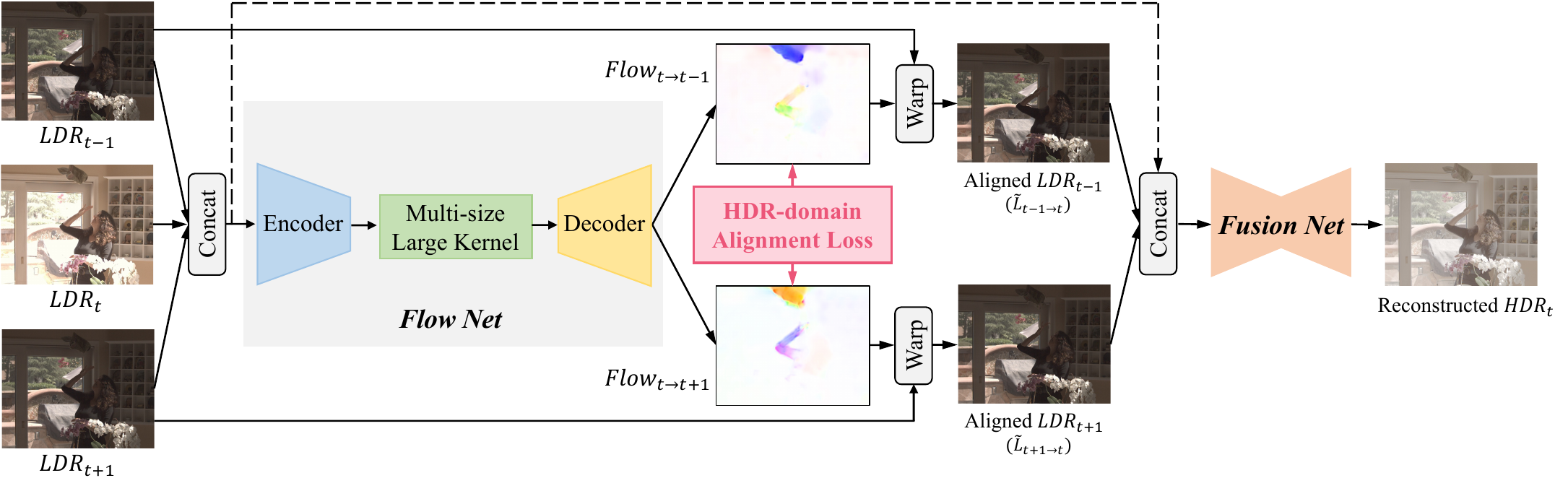}
    \caption{Network architecture of the proposed \abbrmethod .
    We first estimate bidirectional optical flows through the proposed flow network. Then, we align the neighboring frames to the reference frame $t$ based on these estimated flows. To achieve accurate alignment, we introduce a novel HDR-domain alignment loss to supervise our flow network. Finally, the aligned frames and the original frames are fused together through the fusion network to reconstruct a high-quality and ghost-free HDR image for the reference frame.
    }
    \label{fig:network}
    \vspace{-10pt}
\end{figure*}

\paragraph{HDR image reconstruction.}
The most popular way for HDR imaging is to merge multi-exposure LDR images, which is similar to HDR video reconstruction from alternating-exposure LDR frames. Early works~\cite{sen2012robust,hu2013hdr,oh2014robust,ma2017robust} utilize image alignment to reduce ghosting artifacts in dynamic scenes. With the rise of deep neural networks, many works~\cite{wu2018deep,yan2019attention, yan2022dual} directly learn the complicated mapping between LDR and HDR using CNNs. Wu \etal~\cite{wu2018deep} formulated HDR imaging as an image translation problem and proposed the first non-flow-based network for HDR imaging. Yan \etal~\cite{yan2019attention, yan2022dual} proposed a spatial attention module to suppress undesired content from non-reference images and then used a non-local network to merge them. Based on this spatial attention, several methods~\cite{chen2022attention,liu2021adnet,liu2022ghost,liu2023joint} have been proposed to remove ghosting artifacts. Unluckily, spatial attention produces unsatisfactory results when motion occurs in over-exposed regions or under-exposed regions. To mitigate this, 
Yan \etal~\cite{yan2023unified} proposed to integrate similar content from other non-reference images by patch aggregation. However, these methods rely on a fixed-exposure reference frame, such as medium exposure, making it challenging to apply them to HDR video reconstruction with alternating-exposure reference frames.

\vspace{2mm}
\noindent\textbf{HDR video reconstruction.}
Some existing methods rely on dedicated hardware solutions for direct HDR video acquisition, such as scanline exposure/ISO~\cite{heide2014flexisp,choi2017reconstructing} and internal/external beam splitter~\cite{tocci2011versatile, kronander2014unified,mcguire2007optical}. However, these methods require sophisticated designs and are costly.

A practical solution for HDR video reconstruction is to merge multiple LDR images with varying exposures. Kang \etal~\cite{kang2003} introduced the first algorithm of this category by first aligning neighboring frames to the reference frame using global and local registration and then merging the aligned images to an HDR image. Mangiat \etal~\cite{mangiat2010high} improved this method by block-based motion estimation and refined the motion vectors. Kalantari \etal~\cite{kalantari13} adopted patch-based optimization to reconstruct missing images with different exposures. Recently, Kalantari \etal~\cite{kalantari19} presented the first end-to-end CNN-base framework that consists of a flow network for alignment and a weight network for merging the aligned LDR images. Follow this, Chen \etal~\cite{chen2021hdr} performed a more sophisticated alignment by incorporating deformable convolution~\cite{deformable} after the coarse alignment using optical flow. Chung \etal~\cite{lan_hdr} aligned adjacent frames to the reference frame by computing luminance-based attention score. However, these methods suffer from flow estimation errors and attention calculation errors when large motions occur. Different from them, we propose a robust HDR-oriented flow estimator to achieve precise alignment. Meanwhile, our method is significantly faster than theirs.

\section{HDRFlow}
\label{sec:method}

\begin{figure*}
    \centering
    \includegraphics[width=0.75\linewidth]{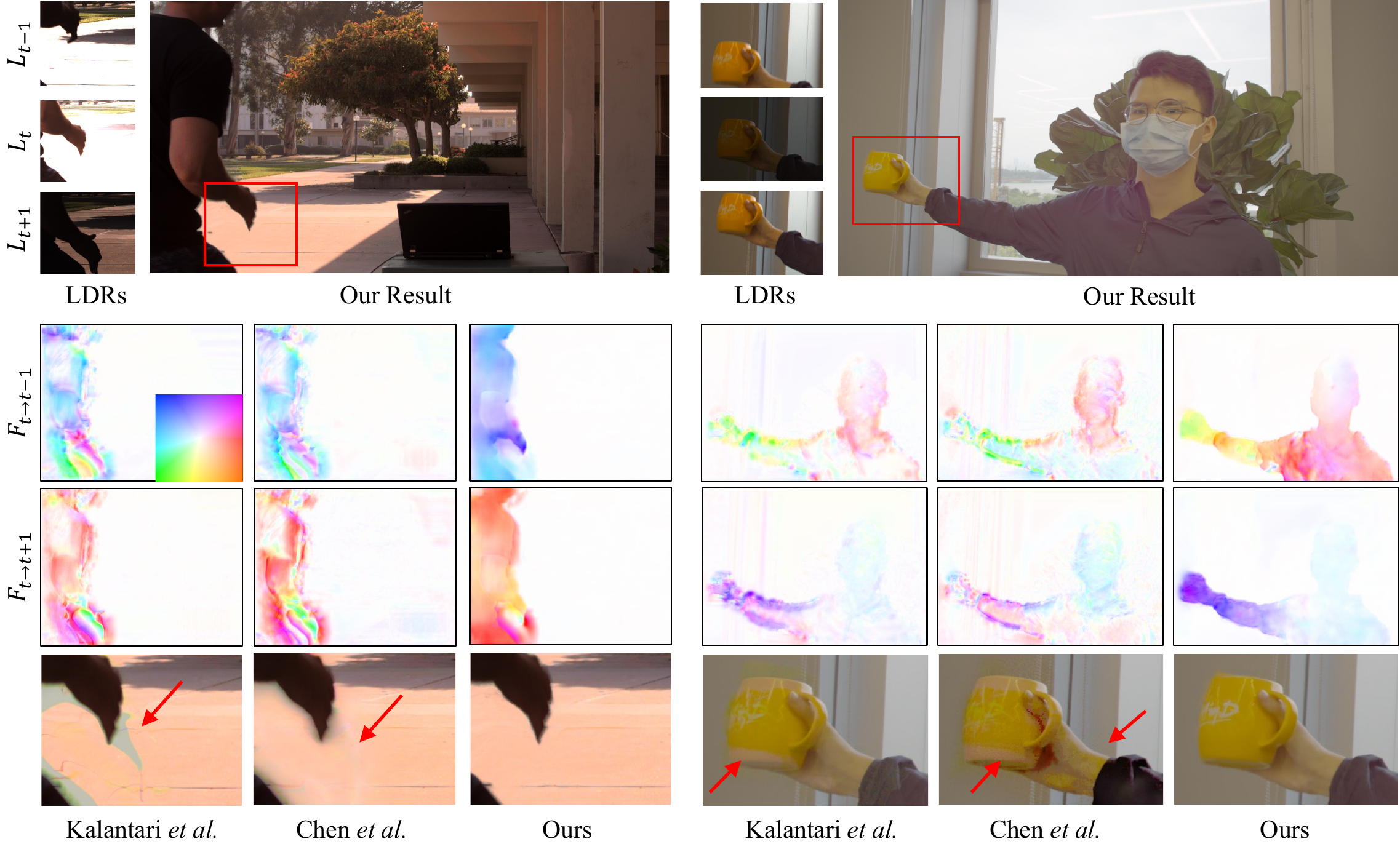}
    \caption{Qualitative comparisons with flow-based methods. Left: 3-Exposure scene from the Kalantari13 dataset~
    \cite{kalantari13}. Right: 2-Exposure scene from the DeepHDRVideo dataset~\cite{chen2021hdr}. Flow visualization is based on the color wheel shown on the corner of the first flow map.}
    \label{fig:comp_flow}
    \vspace{-15pt}
\end{figure*}

In HDR fusion, the input LDR video consists of LDR frames $\{L_t\}$ captured under different exposures $\{e_t\}$, where $t = 1,\dots, n$. Our goal is to efficiently reconstruct high-quality HDR video, consisting of HDR frames $\{H_t\}$. Following the convention~\cite{kalantari19}, we evaluate two different types of input: three frames with two alternating exposures \{EV-3, EV+0, EV-3, \dots\}, and five frames for three alternating exposures \{EV-2, EV+0, EV+2, EV-2, EV+0, \dots\}. In both cases, the middle frame is selected as the reference, and the rest frames are fused to the reference frame to generate HDR output. For simplicity, we introduce our algorithm for handling videos captured with two alternating exposures in this paper, and discuss the extension to three exposures in the \textbf{supplementary material}.

The overview of the proposed network framework is illustrated in \cref{fig:network}. Our framework is composed of the flow network and the fusion network. Firstly, we input three LDR frames into the flow network to estimate bidirectional optical flows, denoted as $F_{t\rightarrow{t-1}}$ and $F_{t\rightarrow{t+1}}$. These flows are utilized to warp the two neighboring frames to the reference frame. Subsequently, we feed the following information to the fusion network: the warped neighboring frames, the reference frame, the original neighboring frames in the LDR domain, and their counterparts in the linear HDR domain.
The mapping function that takes the LDR frame $L_t$ to the linear HDR domain ${I}_{t}$ is defined as:
\begin{align}
    \label{eq:ldr_to_hdr}
    {I}_{t} = L_{t}^{\gamma}/e_t,
\end{align}
where $e_t$ is the exposure time of $L_t$ and $\gamma=2.2$. The fusion network estimates the blending weights, which are then utilized to fuse the linear HDR frames into the final HDR output.

\subsection{Flow Network with Multi-size Large Kernel}
To reconstruct the missing content at frame $t$, aligning neighboring frames with the reference frame is required. To accomplish this, we design an efficient flow network tailored for HDR fusion task to estimate the flow field from the reference frame $t$ to the neighboring frames, that is, previous frame $t-1$ and next frame $t+1$.

\vspace{2mm}
\noindent\textbf{Network design.} The flow network consists of encoder, multi-size large kernel and decoder. Specifically, we incorporate multi-size large kernel at the end of the encoder, which can effectively model large motions at a negligible computational cost. In contrast to previous works~\cite{kalantari19,chen2021hdr}, our network is a single feed-forward network that does not require any iterative prediction and estimated intermediate warped frames. Thus, our flow network is efficient and can predict bidirectional optical flows within a 10ms runtime for 720p resolution. Flow network details can be found in the \textbf{supplementary materials}.

Due to the exposure difference between input frames, we adjust the exposure of the reference frame $t$ to match neighboring frames before injecting it into the flow network, 
\begin{align}
    \label{eq:2}
    {g}_{t+1}(L_t) = \text{clip}\Big( ((L_{t}^{\gamma}/e_t)e_{t+1})^{1/\gamma} \Big),
\end{align}
where ${g}_{t+1}(L_t)$ is adjusted reference frame. We concatenate $L_{t-1}$,  ${g}_{t+1}(L_t)$ and $L_{t+1}$ and send them into the flow network. The encoder of the flow network consists of two subnetworks, one builds a feature pyramid and another one builds an image pyramid.
The feature pyramid consists of 8 residual blocks, 2 at 1/2 resolution, 2 at 1/4 resolution, 2 at 1/8 resolution, and 2 at 1/16 resolution. 
The image pyramid is obtained by applying pooling operations on the concatenated LDR frames. We concatenate the feature pyramid and the image pyramid at 1/4, 1/8, and 1/16 resolution. Finally, we obtain the flow feature $Z_{e}$ ($Z_{e} \in \mathbb{R}^{H/16 \times W/16 \times 256}$). $H$ and $W$ represent image height and width. At the end of the encoder, we perform the proposed multi-size large kernel to increase the receptive field and model large motions. 

The decoder of the flow network consists of two upsampling blocks and a flow head. 
After upsampling the flow feature to 1/4 resolution, the flow head is applied to predict the bidirectional optical flows. 
The predicted flows are at 1/4 resolution, and we upsample them to full resolution by bilinear interpolation. The final bidirectional flows are denoted as ${F}_{t\rightarrow{t-1}}$ and ${F}_{t\rightarrow{t+1}}$. The two neighboring frames can then be aligned to the reference frame as \{$\tilde{L}_{t-1\rightarrow{t}}$, $\tilde{L}_{t+1\rightarrow{t}}$\}.

\vspace{2mm}
\noindent\textbf{Multi-size Large Kernel.} In recent literature~\cite{ding2022scaling, largekernel3d}, it has been observed that large kernel convolutions have much larger effective receptive fields, leading to improved performance in segmentation and detection areas. Inspired by this, we designed the multi-size large kernel convolutions to increase the receptive field and model large motions. As shown in \cref{fig:network}, we place the multi-size large kernel at the coarsest resolution of the flow network to minimize computational costs. The multi-size large kernel consists of three different-sized large kernel convolutions (\ie, $7\!\times\!7$, $9\!\times\!9$, and $11\!\times\!11$), each modeling different degrees of large motions,
\begin{equation} 
\begin{aligned}
{Z}_{mlk} = \text{Concat}\{\text{DConv}_{7\times7}({Z}_{e}),
    \text{DConv}_{9\times9}({Z}_{e}), \\ \text{DConv}_{11\times11}({Z}_{e})\},
\end{aligned}
\end{equation}
where $\text{DConv}$ denotes depth-wise convolution. Then, we apply a $1\times1$ convolution to merge the concatenated feature ${Z}_{mlk}$, and add the merged feature to ${Z}_{e}$.

\begin{figure}
\centering
{\includegraphics[width=0.9\linewidth]{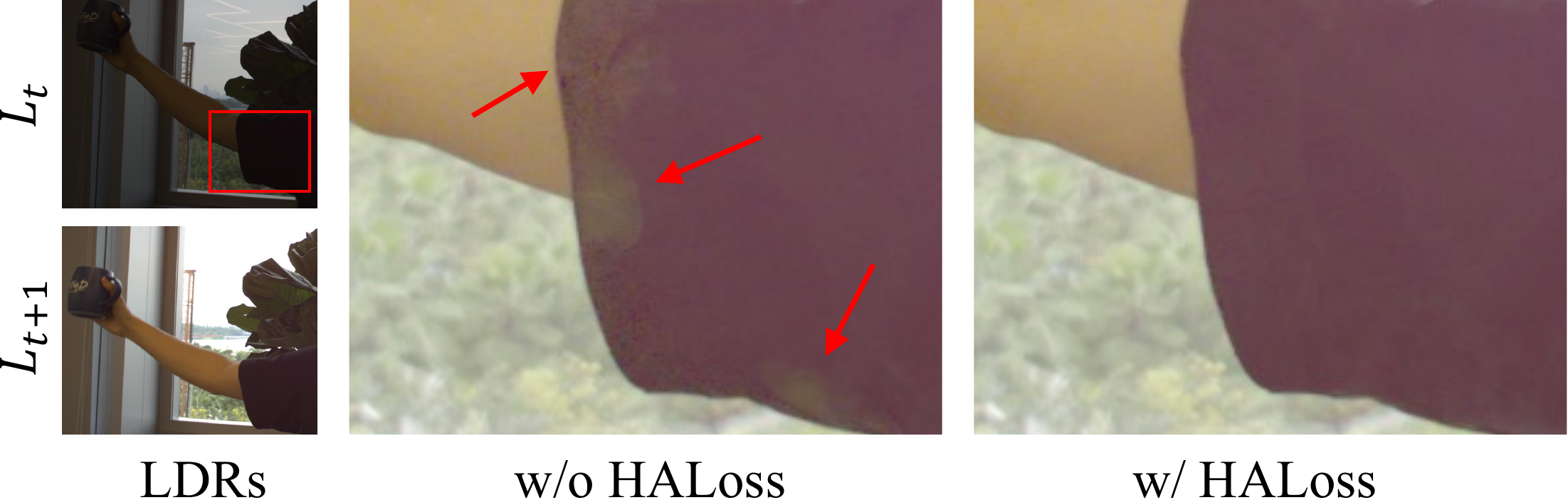}}
\caption{Effectiveness of HALoss.}\label{fig:haloss}
\vspace{-15pt}
\end{figure}

\subsection{HDR-domain Alignment Loss}
To reconstruct a high-quality, artifact-free HDR video, accurate alignment between the reference frame and neighboring frames is crucial. 
However, previous works~\cite{kalantari19,chen2021hdr,lan_hdr} solely compute the loss between the final estimated and ground truth HDR frame, without direct supervision of the intermediate alignment. Therefore, it is challenging to achieve accurate alignment in over-exposed or under-exposed regions (see \cref{fig:haloss} and \cref{fig:comp_flow}). 

Thus, additional supervision on intermediate alignment is important. An intuitive solution is to use photometric consistency loss between the reference frame and the warped neighboring frames. However, the input LDR frames have different brightness, which violates the photometric consistency assumption.

To deal with brightness differences, we propose the HDR-domain Alignment Loss (HALoss) that is robust to brightness change. Since the training LDR frames were generated from clean HDR videos, where there are no brightness changes, we can use HDR frames to calculate photometric consistency loss. Specifically, our flow network predicts the flow field based on the input LDR frames, but the photometric consistency loss is calculated on the warped clean HDR frames, where the predicted flow fields are used for warping. For better perceptual quality, we compute the loss in the tonemapped HDR space. Following previous works~\cite{kalantari19,yan2019attention,wu2018deep}, we use the differentiable $\mu$-law function as the tonemapping function $\mathcal{T}$:
\begin{equation} 
\mathcal{T}(H) = \frac{\log(1+\mu H)}{\log(1+H)},
\end{equation}
where $\mu$ is set to 5,000. 

Given HDR frames $H_{t-1}, H_t, H_{t+1}$ corresponding to LDR frames $L_{t-1}, L_t, L_{t+1}$, as well as estimated flows $F_{t\rightarrow{t-1}}$ and $F_{t\rightarrow{t+1}}$, the HALoss $\mathcal{L}_{HA}$ is expressed as:
\begin{equation} 
\begin{aligned}
\mathcal{L}^{photo}_{t,t-1} = & \;\parallel \mathcal{T}(H_t) - \mathcal{W}(\mathcal{T}(H_{t-1}), F_{t\rightarrow{t-1}}) \parallel_1,\\
\mathcal{L}^{photo}_{t,t+1} = & \;\parallel \mathcal{T}(H_t) - \mathcal{W}(\mathcal{T}(H_{t+1}), F_{t\rightarrow{t+1}}) \parallel_1,\\
\mathcal{L}_{HA} = & \; (1-M_t) \odot (\mathcal{L}^{photo}_{t,t-1} + \mathcal{L}^{photo}_{t,t+1}),
\end{aligned}
\end{equation}
where $\mathcal{W}(\cdot,\cdot)$ denotes the warping of neighboring frames to the reference frame using optical flow. $\mathcal{L}^{photo}_{t,t-1}$ and $\mathcal{L}^{photo}_{t,t+1}$ denote photometric loss. The $M_t$ is a mask indicating the well-exposed regions of the reference frame $t$. We convert $L_t$ to the YCbCr space to obtain the luminance channel, $Y$. Then, the $M_t$ is defined as $\delta_{low}\!\textless\! Y \!\textless\! \delta_{high}$. $\delta_{low}$  and $\delta_{high}$ respectively denote the low and high luminance thresholds. 
Since our objective is to integrate neighboring frames' information into the reference frame in the not well-exposed regions, we only calculate the HALoss in these regions to learn the HDR-oriented optical flows. As illustrated in \cref{fig:haloss}, our HALoss is effective.

\subsection{Fusion Network}
The objective of the fusion network is to generate a high-quality HDR frame from the reference frame, aligned neighboring frames, and original neighboring frames. In the reference LDR frame, both static regions and moving objects can experience over-exposed or under-exposed conditions. Consequently, aligned neighboring frames (\ie, $\tilde{L}_{t-1\rightarrow{t}}$ and $\tilde{L}_{t+1\rightarrow{t}}$, see \cref{fig:network}) obtained by warping operation primarily contribute to missing content in dynamic regions, while original neighboring frames (\ie, $L_{t-1}$ and $L_{t+1}$) mainly provide missing content in static regions. 

With this consideration, five LDR frames, along with their corresponding linear HDR domain frames (\cref{eq:ldr_to_hdr})
are fed into the fusion network. The fusion network adopts a U-Net architecture with skip connections, comprising three downsampling blocks and three upsampling blocks. Fusion network details can be found in the \textbf{supplementary materials}.
The fusion network outputs the fusion weights for five linear HDR frames. The final HDR frame $\hat{H}_t$ is computed as a weighted average of the five linear HDR frames using their fusion weights as:
\begin{equation} 
\begin{aligned}
\hat{H}_t \!=\! \frac{w_0I_t\!+\!w_1\tilde{I}_{t-1\rightarrow{t}}\!+\!w_2\tilde{I}_{t+1\rightarrow{t}}\!+\!w_3I_{t-1}\!+\!w_4I_{t+1}}{\sum_{j= 0}^{4}w_j}.
\end{aligned}
\end{equation}

\begin{table*}
\setlength{\tabcolsep}{4.5pt}
\centering 
\begin{tabular}{l|cccc|cccc}
\toprule
\multirow{2}{*}{Methods} & \multicolumn{4}{c|}{2-Exposure} & \multicolumn{4}{c}{3-Exposure} \\   
& {PSNR\textsubscript{$T$}} & {SSIM\textsubscript{$T$}} &{HDR-VDP-2} & Time (ms) &{PSNR\textsubscript{$T$}} & {SSIM\textsubscript{$T$}} & {HDR-VDP-2} & Time (ms)\\ 
\midrule
Kalantari13~\cite{kalantari13}    & 37.51   & 0.9016 & 60.16  & - & 30.36   & 0.8133 & 57.68 & - \\
Kalantari19~\cite{kalantari19} & 37.06   & 0.9053 & 70.82  & 230 & 33.21   & 0.8402  & 62.44 & 260  \\
Yan19 ~\cite{yan2019attention} & 31.65   & 0.8757  & 69.05  & 460 & 34.22   & 0.8604 & \underline{66.18} & - \\
Prabhakar~\cite{prabhakar}   & 34.72   & 0.8761  & 68.82 & -   & 34.02   & 0.8633   & 65.00 & -  \\
Chen~\cite{chen2021hdr}     & 35.65   & 0.8949  & \textbf{72.09}  & 550  & 34.15   & 0.8847 & \textbf{66.81} & 570  \\
LAN-HDR~\cite{lan_hdr}   & 38.22  & 0.9100 & 69.15  & 707   & 35.07   & 0.8695 & 65.42 & 905  \\
\midrule
Ours (Vimeo) & \underline{39.20} & \underline{0.9154} & 70.98 & 55 & \underline{36.55} & \underline{0.9039} & 65.89 & 76 \\
Ours (Vimeo+Sintel) & \textbf{39.30} & \textbf{0.9156} & \underline{71.05} & 55 & \textbf{36.65} & \textbf{0.9055} & 66.02 & 76 \\
\bottomrule
\end{tabular}
\caption{Quantitative comparisons of our method with other state-of-the-art methods on the Cinematic Video dataset~\cite{cinematic}. The time is the inference time for the $1920\times1080$ resolution dataset. \textbf{Bold}: best, \underline{underline}: second best.}
\label{tab:synthetic}
\end{table*}

\begin{table*}
\setlength{\tabcolsep}{4.5pt}
\centering 
\begin{tabular}{l|cccc|cccc}
\toprule
\multirow{2}{*}{Methods} & \multicolumn{4}{c|}{2-Exposure} & \multicolumn{4}{c}{3-Exposure} \\  
& {PSNR\textsubscript{$T$}} & {SSIM\textsubscript{$T$}} &{HDR-VDP-2} &Time (ms) &{PSNR\textsubscript{$T$}} & {SSIM\textsubscript{$T$}} & {HDR-VDP-2} &Time (ms) \\ 
\midrule
Kalantari13~\cite{kalantari13} & 40.33   & 0.9409 & 66.11   &- & 38.45   & 0.9489 & 57.31 &- \\
Kalantari19~\cite{kalantari19} & 39.91   & 0.9329 & 71.11  & 200 & 38.78   & 0.9331  & 65.73 & 220 \\
Yan19~\cite{yan2019attention} & 40.54   & 0.9452  & 69.67 &280  & 40.20   & 0.9531 & 68.23 & - \\
Prabhakar~\cite{prabhakar}   & 40.21   & 0.9414  & 70.27   &- & 39.48   & 0.9453   & 65.93 & - \\
Chen~\cite{chen2021hdr}     & 42.48   & \textbf{0.9620}  & 74.80  &522  & 39.44   & \textbf{0.9569} & 67.76 & 540 \\
LAN-HDR~\cite{lan_hdr}   & 41.59  & 0.9472 & 71.34   &415  & 40.48   & 0.9504 & 68.61 &525  \\
\midrule
Ours (Vimeo) & \underline{43.18} & 0.9510 & \underline{77.11} & 35  & \underline{40.45} & 0.9530 & \underline{72.30} &50 \\
Ours (Vimeo+Sintel) & \textbf{43.25} & \underline{0.9520} & \textbf{77.29} &35  & \textbf{40.56} & \underline{0.9535} & \textbf{72.42} &50 \\
\bottomrule
\end{tabular}
\caption{Quantitative comparisons of our method with other state-of-the-art methods on the DeepHDRVideo dataset. The time is the inference time for the $1536\times813$ resolution dataset.}
\label{tab:deephdrvideo}
\vspace{-10pt}
\end{table*}

\subsection{Training Scheme and Loss}

\noindent\textbf{New training scheme.} Existing video HDR networks~\cite{chen2021hdr,lan_hdr} are mostly trained on real videos, such as Vimeo-90K, which often lack instances of large motion. Synthetic data, such as Sintel~\cite{sintel}, include examples of large motion. In this paper, we propose to incorporate both synthetic and real videos for training. Specially, the Sintel data provides ground-truth forward flow ($F_{t\rightarrow{t+1}}^{gt}$), and we generate backward flow ($F_{t\rightarrow{t-1}}^{gt}$) using the pre-trained RAFT~\cite{raft} flow network. Then, we use the forward and backward flow supervision to further enhance the accuracy and robustness of optical flow.

\vspace{2mm}
\noindent\textbf{Total loss.}
The total loss consists of reconstruction loss, alignment loss, and flow loss. We compute the reconstruction loss $\mathcal{L}_{rec}$ between the predicted ${\hat{H}}_t$ and ground-truth ${H}_t^{gt}$ using $\mathcal{L}_1$ loss:
\begin{equation} 
\begin{aligned}
\mathcal{L}_{rec} =  \parallel \mathcal{T}({\hat{H}}_t) - \mathcal{T}({H}_t^{gt})) \parallel_1.
\end{aligned}
\end{equation}
For Sintel data, we additionally compute the loss between predicted flow and ground truth flow to further improve the flow’s robustness against large motions. The flow loss $\mathcal{L}_{flow}$ is defined as:
\begin{equation} 
\begin{aligned}
\mathcal{L}_{flow} =  \parallel F_{t\rightarrow{t-1}}\!-\!F_{t\rightarrow{t-1}}^{gt} \parallel_1 \!+\! \parallel F_{t\rightarrow{t+1}}\!-\!F_{t\rightarrow{t+1}}^{gt} \parallel_1.
\end{aligned}
\end{equation}
The total loss $\mathcal{L}_{total}$ is represented as:
\begin{equation} 
\begin{aligned}
\mathcal{L}_{total} = \lambda_{1}\mathcal{L}_{rec}+\lambda_{2}\mathcal{L}_{HA}+\lambda_{3}\mathcal{L}_{flow},
\end{aligned}
\end{equation}
where $\lambda_1=1$, $\lambda_2=0.5$, and $\lambda_3=0.001$.

\section{Experiments}
\label{sec:experiment}

\begin{table*}
  \centering
  \begin{tabular}{l|cccc|cc|cc|c}
    \toprule
    \multirow{2}{*}{Model} & \multirow{2}{*}{FN} & \multirow{2}{*}{MLK} & \multirow{2}{*}{HALoss} & Mask for &\multicolumn{2}{c|}{DeepHDRVideo-D} &\multicolumn{2}{c|}{Cinematic Video} & Time\\ 
    &&&& HALoss & {PSNR\textsubscript{$T$}} & {SSIM\textsubscript{$T$}} & {PSNR\textsubscript{$T$}} & {SSIM\textsubscript{$T$}} & (ms)\\
    \midrule
    Base (SPyNet) & & & & & 44.82 & 0.9650 & 38.78 & 0.9137 & 73 \\
    \midrule
    FN &\checkmark & & & & 45.10 & 0.9655 & 38.92 & 0.9133 & 34 \\
    FN+MLK &\checkmark &\checkmark & & & 45.30 & 0.9661 & 39.08 & 0.9142 & 35 \\
    FN+MLK+HA &\checkmark &\checkmark &\checkmark & & 45.41 & 0.9669 & 39.21 & 0.9148 & 35 \\
    Full model (HDRFlow) &\checkmark &\checkmark &\checkmark& \checkmark & \textbf{45.50} & \textbf{0.9683} & \textbf{39.30} & \textbf{0.9156} & 35 \\
    \bottomrule
  \end{tabular}
  \vspace{-2mm}
  \caption{
  Ablation study of \abbrmethod on the dynamic scenes of DeepHDRVideo dataset and Cinematic Video dataset. FN denotes our flow network, and HALoss denotes HDR-domain alignment loss. Following the previous works~\cite{kalantari19,chen2021hdr}, the Base (SPyNet) utilizes SPyNet~\cite{spynet} as the optical flow estimator and does not incorporate HALoss. The time is the inference time for 1536$\times$813 resolution. \textbf{Bold}: Best.}
  \label{tab:ablation}
  \vspace{-15pt}
\end{table*}

\subsection{Experimental Setup}
\textbf{Datasets.} We utilize the high-quality Vimeo-90K~\cite{xue2019video} and Sintel~\cite{sintel} datasets as our training sets. As the Vimeo-90K and Sintel datasets are not tailored for HDR video reconstruction, we convert the original data to LDR sequences with alternating exposures following the previous work\cite{kalantari19,chen2021hdr}.  
During training, we first apply random horizontal/vertical flipping and rotation, and then randomly crop the resulting images to obtain patches of size $256\times256$ as inputs to the network. We evaluate our method on two synthetic videos (POKER FULLSHOT and CAROUSEL FIREWORKS) from the Cinematic Video dataset~\cite{cinematic} and DeepHDRVideo dataset~\cite{chen2021hdr}. The resolution of Cinematic Video dataset is $1920\times1080$, and the resolution of DeepHDRVideo dataset is $1536\times813$. The DeepHDRVideo dataset consists of both real-world dynamic scenes and static scenes that have been augmented with random global motion. We also utilize the HDRVideo dataset~\cite{kalantari13}, which has a resolution of $1280\times720$, for qualitative evaluation.

\vspace{2mm}
\noindent\textbf{Implementation details.} We implement our approach with PyTorch and perform our experiments using an NVIDIA 3090 GPU. We adopt AdamW optimizer~\cite{adam} with $\beta_1=0.9$ and $\beta_2=0.999$. We train our network with 40 epochs using a batch size of 16. The learning rate was initially set to 0.0001 and halved after epochs 20 and 30. In our experiments, we empirically set $\delta_{low}$ to 0.2 and $\delta_{high}$ to 0.8.

\noindent\textbf{Evaluation metrics} We adopt $\text{PSNR}_T$, $\text{SSIM}_T$ and HDR-VDP-2~\cite{hdr_vap_2} as the evaluation metris. $\text{PSNR}_T$, $\text{SSIM}_T$ are computed in the $\mu$-law tonemapped domain.

\subsection{Comparisons with State-of-the-art}

\noindent\textbf{Quantitative evaluations.} \cref{tab:synthetic} and \cref{tab:deephdrvideo} show our quantitative results on the Cinematic Video~\cite{cinematic} and DeepHDRVideo~\cite{chen2021hdr} datasets. Ours (Vimeo) denotes that we only use Vimeo-90K~\cite{xue2019video} as our training dataset, which is consistent with other methods. Ours (Vimeo+Sintel) denotes that we use both Vimeo-90K and Sintel~\cite{sintel} as our training datasets. The performance can be further improved by incorporating Sintel into the training dataset. Our method achieves superior or comparable results to state-of-the-art methods. Specifically, on the Cinematic Video dataset, our method outperforms the second-best method by up to 1.08dB and 1.58dB in terms of PSNR\textsubscript{$T$} for the 2-Exposure and 3-Exposure cases, respectively. On the DeepHDRVideo dataset, our method has also achieved the best results among all methods in terms of PSNR\textsubscript{$T$} and HDR-VDP-2.

\begin{figure*}
\centering
{\includegraphics[width=0.75\linewidth]{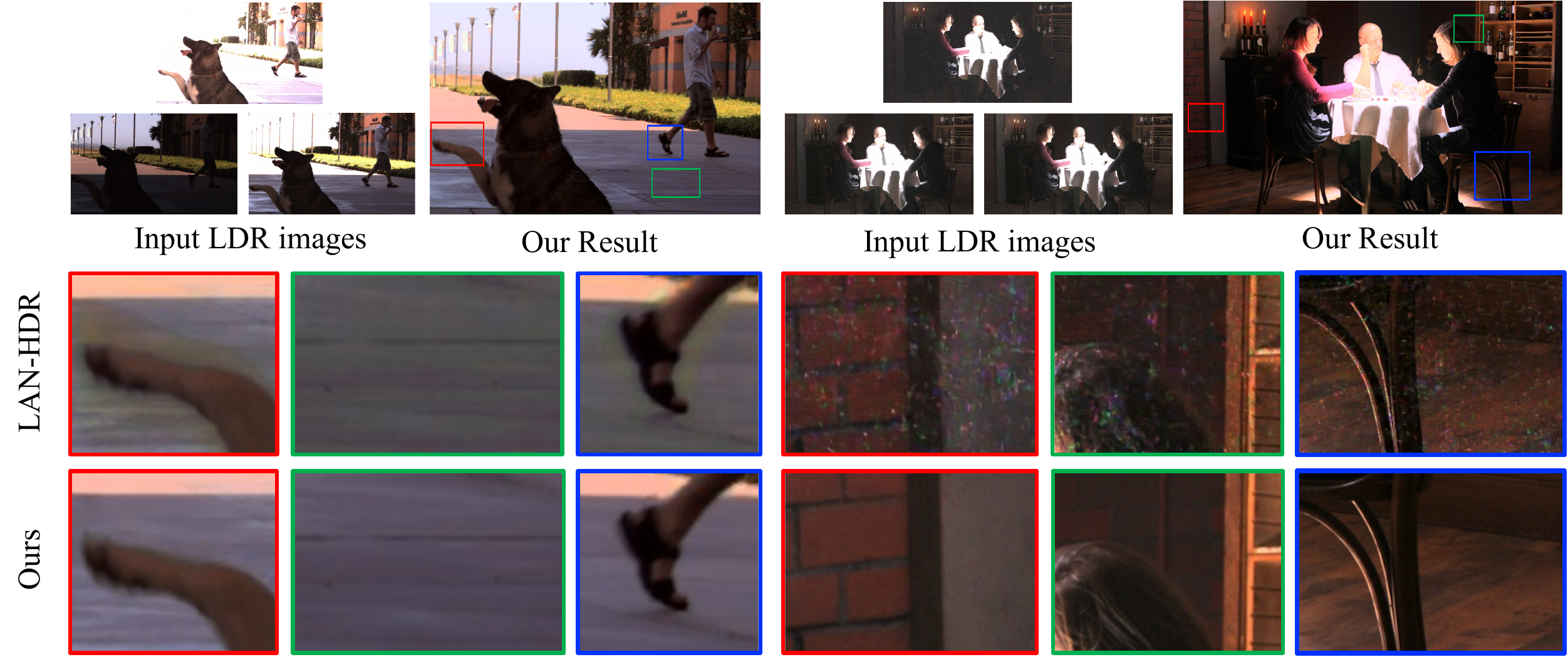}}
\caption{Qualitative comparisons with attention-based method~\cite{lan_hdr}.}\label{fig:comp_att}
\vspace{-20pt}
\end{figure*}

\vspace{2mm}
\noindent\textbf{Qualitative evaluations.} 
We compared the visual results of our approach with flow-based methods~\cite{kalantari19,chen2021hdr} in \cref{fig:comp_flow}, and with attention-based methods~\cite{lan_hdr} in \cref{fig:comp_att}. As shown in \cref{fig:comp_flow},
the optical flows predicted by the methods of Kalantari \etal and Chen \etal are relatively discontinuous, lacking smoothness and completeness. Consequently, their methods exhibit ghosting artifacts in regions with large motions and lose details in saturated regions. In contrast, our predicted optical flow is more accurate and smooth, enabling precise alignment in regions with large motions. Without estimating optical flow, LAN-HDR~\cite{lan_hdr} performs alignment using an attention module. However, using attention easily leads to the aggregation of irrelevant information, resulting in local inconsistency. As shown in \cref{fig:comp_att}, LAN-HDR~\cite{lan_hdr} introduces color distortion due to the aggregation of grass content onto the ground. Additionally, LAN-HDR struggles to handle noise in extremely dark areas. On the contrary, our method can produce pleasing results in these regions.

\vspace{2mm}
\noindent\textbf{Inference time.} To demonstrate the high efficiency of our method, we compare the inference time of our method with other HDR video reconstruction methods in \cref{tab:synthetic} and \cref{tab:deephdrvideo}. For fair comparisons, the inference times of all methods are tested on a single NVIDIA 3090 GPU. Our method is approximately $10\times$ faster than both Chen~\cite{chen2021hdr} and LAN-HDR~\cite{lan_hdr}. To the best of our knowledge, our method is the first real-time HDR video reconstruction method for video sequences captured with alternating exposures.

\subsection{Ablation Study}
We conduct ablation studies to validate the effectiveness of the proposed components with the example of sequences having two alternating exposures. All the quantitative evaluations are conducted on the Cinematic Video dataset and the dynamic scenes of DeepHDRVideo dataset, which contains large motions. We take Base (SPyNet) as the baseline. The Base (SPyNet) uses SPyNet~\cite{spynet} as flow estimator, and the fusion network is consistent with ours. SPyNet~\cite{spynet} utilizes a hierarchical coarse-to-fine architecture that needs iterative prediction and estimated intermediate warped frames. This process is computationally intensive and cannot achieve real-time performance. Furthermore, saturated and dark regions in LDR images lack texture information to the extent that at finer resolutions, SPyNet~\cite{spynet} struggles to accurately estimate optical flow relying solely on image information.
In contrast, our flow network (FN) efficiently leverages multi-resolution feature information to estimate accurate optical flow.
Compared to SPyNet (\cref{tab:ablation}), our FN obviously improves performance and greatly reduces inference time. When incorporating the proposed MLK into FN, our model better models large motions, leading to further improvements in the results. 

To achieve precise alignment, we introduce a novel HDR-domain Alignment Loss (HALoss). As shown in \cref{tab:ablation}, HALoss significantly improves performance on benchmarks and does not increase the inference time. Since our objective is to integrate neighboring frames' information into the reference frame in the not-well-exposed regions,
we define a luminance mask that indicates over-exposed and under-exposed regions, and only compute HALoss within these regions to learn the HDR-oriented optical flow. This mask further improves the results on benchmarks. Visual comparisons are shown in \cref{fig:haloss}.

\begin{table} \small
\setlength{\tabcolsep}{2.pt}
\centering 
\begin{tabular}{l|cc|cc|c}
\toprule
\multirow{2}{*}{Method}  & \multicolumn{2}{c|}{DeepHDRVideo-D} & \multicolumn{2}{c}{Cinematic Video} & Time\\ 
 & {PSNR\textsubscript{$T$}} & {SSIM\textsubscript{$T$}} & {PSNR\textsubscript{$T$}} & {SSIM\textsubscript{$T$}}  & (ms) \\ 
\midrule
RAFT+fusion & 44.85 & 0.9634 & 38.63 & 0.9131 & 532\\
Ours & \textbf{45.50} & \textbf{0.9683} & \textbf{39.30} & \textbf{0.9156} & \textbf{35}\\
\bottomrule
\end{tabular}
\caption{Quantitative comparison with RAFT+fusion.}
\label{tab:compare_raft}
\vspace{-15pt}
\end{table}

\subsection{Analysis}
\noindent\textbf{Comparison with state-of-the-art flow method.} We compare our method with the RAFT+fusion method, which is constructed by using a pre-trained RAFT flow network as the flow estimator and employing the same fusion network as in our approach. For training RAFT+fusion, we freeze the optical flow network and only update the fusion network parameters. Quantitative comparisons are shown in \cref{tab:compare_raft}. Our HDRFlow not only outperforms RAFT+fusion but is also about 15$\times$ faster than it. Visual comparisons are shown in \cref{fig:teaser}, RAFT can effectively match visible objects, exhibiting clear flow boundaries, but this flow is incapable of handling occlusions during the alignment. As a result, the fused HDR image exhibits ghosting artifacts in occluded regions. 
Furthermore, RAFT performs poorly in large textureless areas due to limited receptive field.

\begin{figure}
\centering
{\includegraphics[width=0.9\linewidth]{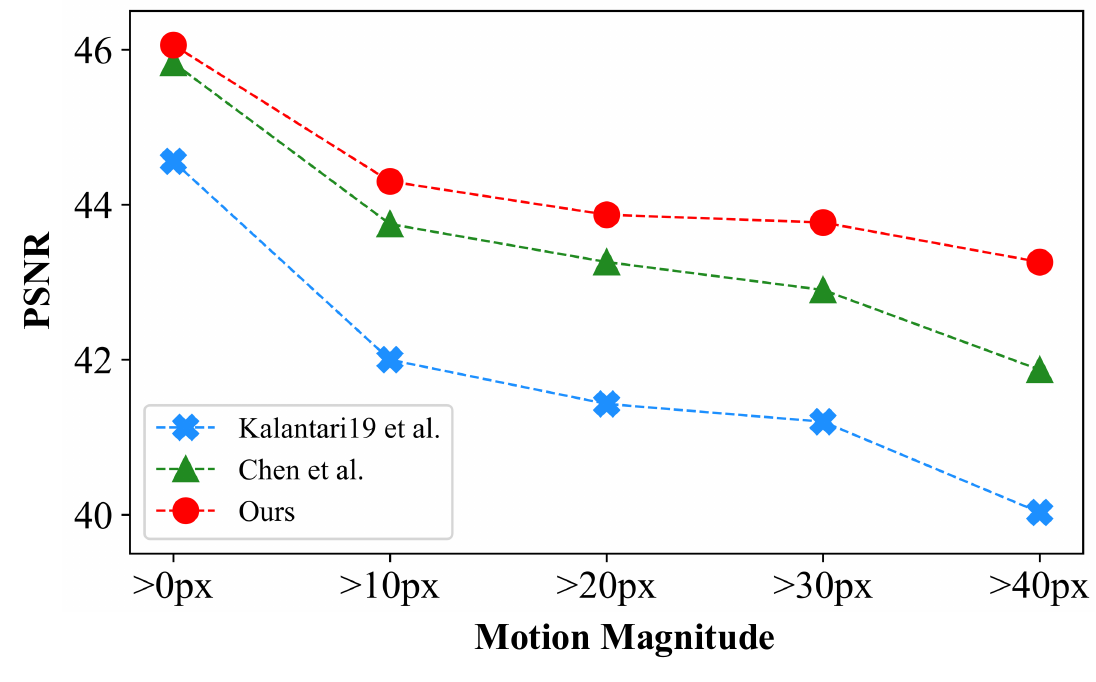}}
\caption{Comparisons with flow-based methods~\cite{kalantari19,chen2021hdr} across different motion magnitude ranges.}
\label{fig:motion}
\vspace{-15pt}
\end{figure}

\vspace{2mm}
\noindent\textbf{Robustness to large motions.} We evaluate the performance of our method with previous flow-based methods~\cite{kalantari19,chen2021hdr} at different motion magnitudes (\cref{fig:motion}). To construct the evaluation dataset, we use RAFT~\cite{raft} to process dynamic scenes of DeepHDRVideo and obtain optical flow maps. Then, we manually crop out these images with reasonable flow predictions. We divide cropped images into $128\times128$ blocks and calculate the average motion magnitude within each block. Finally, we evaluate the PSNR of the blocks corresponding to different motion magnitudes. As shown in \cref{fig:motion}, our HDRFlow is more robust compared to other methods as the motion magnitude increases.

\section{Conclusion}
\label{sec:conclusion}
In this paper, we have proposed a robust and efficient flow estimation algorithm tailored for real-time HDR video reconstruction, named HDRFlow. The HDRFlow has introduced three novel designs: an HDR-domain alignment loss, an efficient flow network with a multi-size large kernel, and a new HDR flow training scheme. Extensive experiments demonstrate that our approach surpasses state-of-the-art methods on public benchmarks. In particular, our approach exhibits superior performance in handling large motion regions. To the best of our knowledge, HDRFlow is the first real-time HDR video reconstruction method.


\noindent\textbf{Acknowledgement.} This work is supported by National Natural Science Foundation of China (62122029, 62061160490, U20B200007). This work is also partially supported by the National Key R\&D Program of China (NO.2022ZD0160101).

{
    \small
    \bibliographystyle{ieeenat_fullname}
    \bibliography{main}
}

\clearpage
\setcounter{page}{1}
\maketitlesupplementary

\begin{figure}
\centering
{\includegraphics[width=1.0\linewidth]{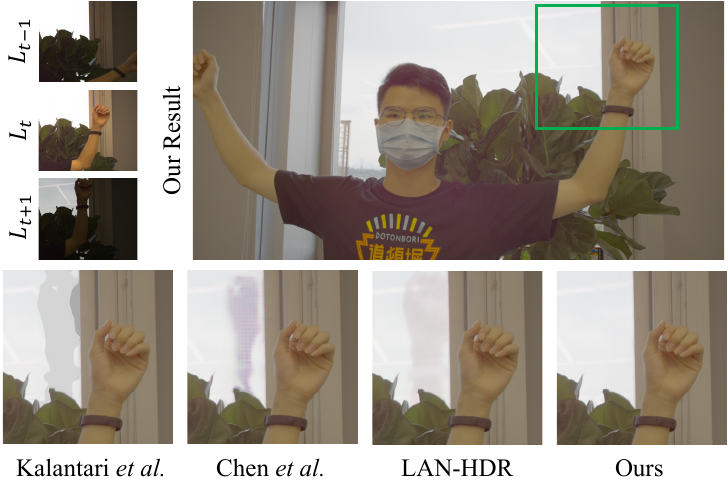}}
\caption{Qualitative comparisons on the DeepHDRVideo dataset (3-Exposure). Compared to previous methods~\cite{kalantari19,chen2021hdr,lan_hdr}, our approach produces ghosting-free results under large motions.}\label{fig:comp_all}
\end{figure}

\section{More Experimental Results}
\label{sec:experiment}
\subsection{More Comparisons with Previous Methods}
\begin{figure*}
    \centering
    \includegraphics[width=1.0\linewidth]{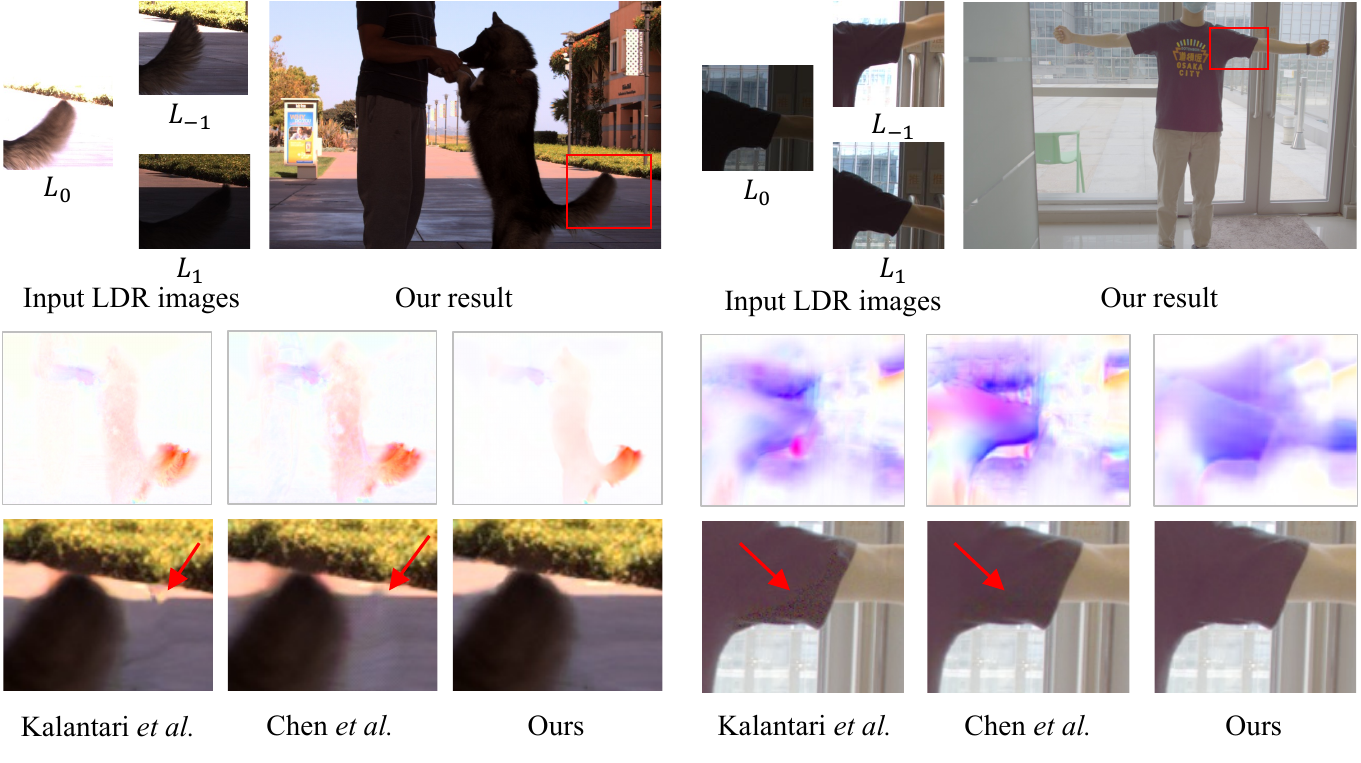}
    \caption{Comparisons with the state-of-the-art methods. The optical flows predicted by the methods of Kalantari \etal~\cite{kalantari19} and Chen \etal~\cite{chen2021hdr} are discontinuous, lacking smoothness and completeness. As a result, their HDR outputs exhibit ghosting artifacts and noise, and lose details in saturated regions. In comparison, our predicted flows are more accurate and smooth, enabling precise alignment in regions with large motions. Thus, our method produces high-quality HDR output.
    }
    \label{fig:comp_sota_flow.}
\end{figure*}

As shown in \cref{fig:comp_all}, we provide more visual comparisons with previous methods, Kalantari19~\cite{kalantari19}, Chen21~\cite{chen2021hdr}, and LAN-HDR~\cite{lan_hdr}. The previous methods struggle to handle large motions, resulting in ghosting artifacts in the final HDR output. In comparison, our method produces high-quality, ghosting-free HDR results. We also provide more visual comparisons with flow-based methods~\cite{kalantari19,chen2021hdr}, shown in \cref{fig:comp_sota_flow.}. The optical flows predicted by the methods of Kalantari \etal~\cite{kalantari19} and Chen \etal~\cite{chen2021hdr} are discontinuous, lacking smoothness and completeness. As a result, their HDR outputs exhibit ghosting artifacts and noise,  and lose details in saturated regions. In comparison, our predicted flows are more accurate and smooth, enabling precise alignment in regions with large motions.

\begin{figure}
\centering
{\includegraphics[width=1.0\linewidth]{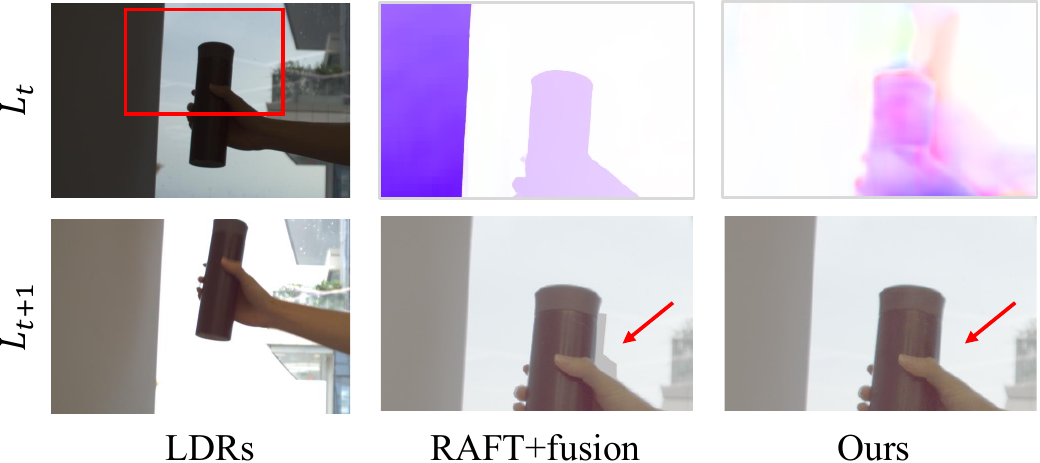}}
\caption{Comparisons with RAFT+fusion. We construct RAFT+fusion by using a pre-trained RAFT~\cite{raft} flow network as the flow estimator and employing the same fusion network as in our approach. RAFT can effectively match visible objects, exhibiting clear flow boundaries, but this flow is incapable of handling occlusions during the alignment. As a result, the fused HDR image exhibits ghosting artifacts in occluded regions. In comparison, our method effectively handles occluded regions by learning an HDR-oriented flow.}\label{fig:comp_raft}
\end{figure}

\cref{fig:comp_raft} shows the comparison between our method and RAFT+fusion. RAFT's~\cite{raft} flow is sub-optimal, and alignment may fail in occluded regions. In contrast, our method effectively handles occluded regions by learning an HDR-oriented flow.

\subsection{Runtime of Each Module}
We benchmark the runtime of each module during inference for the 2-Exposure case. As shown in \cref{tab:module_time}, our Flow Network only takes 10ms and 15ms for resolutions of $1280\times720$ and $1536\times813$, respectively. This is faster than most existing flow methods~\cite{spynet,raft}.

\begin{table}
\centering 
\begin{tabular}{lcc}
\toprule
{Module}  & $1280\times720$ & $1536\times813$ \\  
\midrule
Flow Net with MLK & 10 ms & 15 ms \\
Fusion Net & 15 ms & 20 ms \\
\bottomrule
\end{tabular}
\caption{Runtime time analysis of each module for 2-Exposure case. The input resolutions are $1280\!\times\!720$ and $1536\!\times\!813$, respectively.}
\label{tab:module_time}
\end{table}

\section{Network Details for the Proposed HDRFlow}
\subsection{Details of Flow Network with Multi-size Large Kernel}
\begin{figure*}
    \centering
    \includegraphics[width=1.0\linewidth]{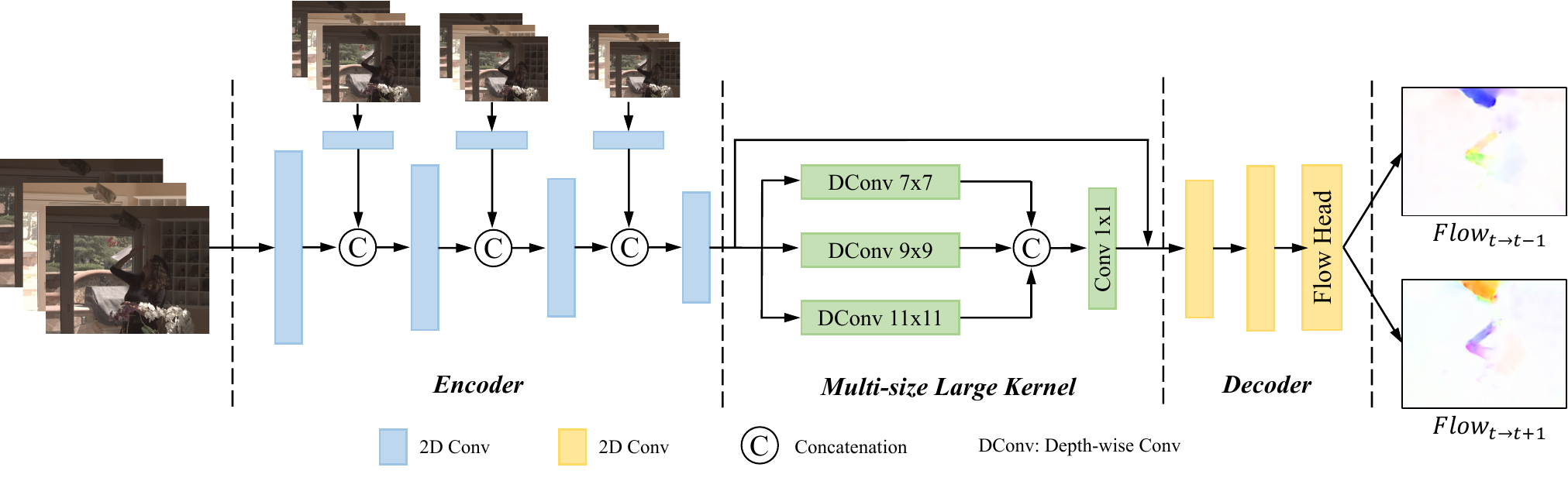}
    \caption{Flow Network with Multi-size Large Kernel. The flow network consists of the encoder, a multi-size large kernel, and a decoder. The flow network takes LDR images as input and outputs bidirectional optical flows.
    }
    \label{fig:flownet}
\end{figure*}

\cref{fig:flownet} shows the architecture of our flow network. The encoder of the flow network consists of two subnetworks, one builds a feature pyramid and another one builds an image pyramid. The feature pyramid consists of 8 residual blocks, 2 at 1/2 resolution, 2 at 1/4 resolution, 2 at 1/8 resolution, and 2 at 1/16 resolution. The corresponding channel numbers are 32, 64, 128, and 256, respectively. The image pyramid is obtained by applying pooling operations on the concatenated LDR frames. We concatenate the feature pyramid and the image pyramid at 1/4, 1/8, and 1/16 resolution. Finally, we obtain the flow feature at the 1/16 resolution. 

Then, we perform the multi-size large kernel convolutions to increase the receptive field and model large motions. The multi-size large kernel consists of three different-sized large kernel convolutions, \ie, $7\times7$, $9\times9$, and $11\times11$, each modeling different degrees of large motions. We utilize depth-wise convolutions, which almost do not increase the computational costs, shown in \cref{fig:flownet}. 

The decoder of the flow network consists of two upsampling blocks and a flow head. Each upsampling block has a $4\times4$ kernel deconvolution with a stride of 2. After each upsampling block, features are concatenated with a skip-connection, and a $1\times1$ convolution followed by a $3\times3$ convolution is applied to merge the skipped and upsampled features for the current resolution. The upsampled flow features are at resolutions of 1/8 and 1/4, with channel numbers of 128 and 64, respectively. After upsampling the flow feature to 1/4 resolution, the flow head is applied to predict the bidirectional optical flows. The flow head consists of three $5\times5$ kernel convolutions.

\subsection{Details of Fusion Network}
The fusion network adopts a U-Net architecture with skip connections, comprising three downsampling blocks and three upsampling blocks. In more detail, each downsampling block consists of a $3\times3$ convolution with a stride of 2, followed by a $3\times3$ convolution with a stride of 1. After three downsampling blocks, we obtain features at three different resolutions: 1/2, 1/4, and 1/8 of the original resolution. The corresponding channel numbers for these resolutions are 32, 64, and 128, respectively. Each upsampling block consists of a $4\times4$ deconvolution with stride 2, followed by a $3\times3$ convolution with stride 1. The fusion network outputs the fusion weights for five LDR frames in the linear domain.

\subsection{Generation of Aligned Neighboring Frames}
We use predicted bidirectional optical flows, $F_{t\rightarrow{t-1}}$ and $F_{t\rightarrow{t+1}}$, to align neighboring frames to reference frame via warping operation,
\begin{equation} 
\begin{aligned}
\tilde{L}_{t-1\rightarrow{t}} = & \; \mathcal{W}(L_{t-1}, F_{t\rightarrow{t-1}}),\\
\tilde{L}_{t+1\rightarrow{t}} = & \; \mathcal{W}(L_{t+1}, F_{t\rightarrow{t+1}}).
\end{aligned}
\end{equation}
The $\tilde{L}_{t-1\rightarrow{t}}$ and $\tilde{L}_{t+1\rightarrow{t}}$ are aligned neighboring frames.

\begin{figure*}
    \centering
    \includegraphics[width=1.0\linewidth]{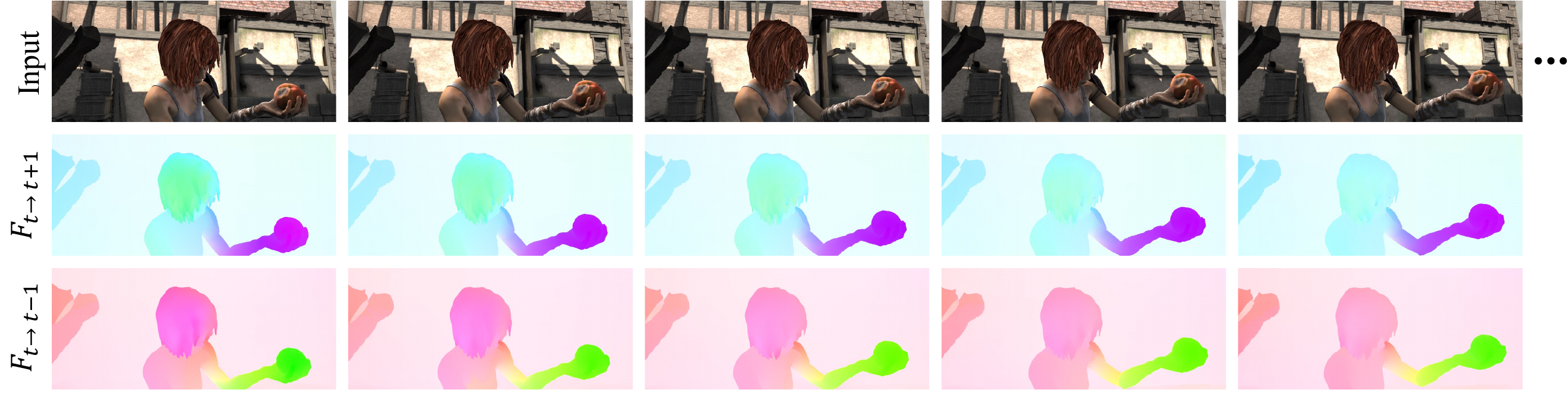}
    \caption{Optical flow labels for Sintel~\cite{sintel} dataset. The first row is video frames of Sintel dataset, and the second row is ground-truth forward optical flow from frame t to t+1. The Sintel does not provide backward flow. Therefore, we use pre-trained RAFT~\cite{raft} flow network to generate backward optical flow from frame t to t-1 as pseudo-labels, shown in the third row. 
    }
    \label{fig:sintel}
\end{figure*}

\subsection{Optical Flow Labels for Sintel}
We use the Sintel dataset as our training dataset. As shown in \cref{fig:sintel}, the Sintel dataset provides ground-truth forward flow ($F_{t\rightarrow{t+1}}$, the second row of \cref{fig:sintel}). However, the Sintel does not provide backward flow. To train our flow network, we use pre-trained RAFT~\cite{raft} flow network to generate backward flow ($F_{t\rightarrow{t-1}}$, the third row of \cref{fig:sintel}) as pseudo-labels.

\subsection{Extension to Three Exposures}
We have illustrated our HDRFlow for handling videos captured with two alternating exposures in the paper. Here we discuss the extension to three exposures.
\paragraph{Review of two-exposure model} For sequences captured with two alternating exposures (\eg, \{EV-3, EV+0, EV-3, \dots\}), our flow network takes three LDR frames \{$L_{t-1}$, $L_{t}$, $L_{t+1}$\} as input and estimates the optical flow, $F_{t\rightarrow{t-1}}$ and $F_{t\rightarrow{t+1}}$. Then, we align the neighboring frames \{$L_{t-1}$, $L_{t+1}$\} to the reference frame $t$ based on these estimated flows. Finally, the aligned frames (2 images) and the original frames (3 images) in the linear domain are fused together through the fusion network to reconstruct a high-quality and ghost-free HDR image for the reference frame.

\paragraph{HDRFlow for sequences with three exposures} For sequences with three alternating exposures (\eg, \{EV-2, EV+0, EV+2, EV-2, EV+0, \dots\}), our HDRFlow takes five frames \{$L_{t-2}$, $L_{t-1}$, $L_{t}$, $L_{t+1}$, $L_{t+2}$\} as input and estimates the HDR image for the reference frame $t$. Specifically, we adjust the exposure of the reference frame $t$ to match neighboring frames before injecting it into the flow network. Thus, the flow network takes \{$L_{t-2}$, $g_{t+1}(L_t)$, $L_{t+1}$\} and \{$L_{t-1}$, $g_{t+2}(L_t)$, $L_{t+2}$\} as input and estimates four flow maps, $F_{t\rightarrow{t-2}}$, $F_{t\rightarrow{t-1}}$, $F_{t\rightarrow{t+1}}$, and $F_{t\rightarrow{t+2}}$. The four neighboring frames can then be aligned to the reference frame as \{$\tilde{L}_{t-2\rightarrow{t}}$, $\tilde{L}_{t-1\rightarrow{t}}$, $\tilde{L}_{t+1\rightarrow{t}}$, $\tilde{L}_{t+2\rightarrow{t}}$\} using the estimated flows. The aligned frames (4 images) and the original input frames (5 images) in both the LDR and linear HDR domain are used as the input (54 channels) for the fusion network to estimate 9 fusion weight maps. Then, the HDR image for the reference frame $t$ can be reconstructed as the weighted average of 9 input images in the linear domain. The overall architectures of both the flow network and fusion network remain consistent between sequences with two and three exposures. The sole distinction lies in the channel numbers at the input and output layers.

\end{document}